\documentclass[
]{ceurart}

\usepackage{listings}
\usepackage[dvipsnames]{xcolor}
\usepackage{todonotes}

\usepackage{amsmath}
\usepackage{amssymb}

\usepackage{graphicx}
\usepackage{xspace}
\usepackage{comment}
\usepackage{booktabs}
\usepackage{multirow}

\usepackage{tikz}
\usetikzlibrary{arrows.meta, positioning}

\newcommand{\Math}[1]{\ensuremath{#1}}

\newcommand{\modecal}[1]{{\Math{\mathcal{#1}}}}

 \newcommand{\D}{\modecal{D}}

\newcommand{\U}{\modecal{U}}

\newcommand{\ltl}{\ensuremath{\textsc{LTL}}\xspace}
\newcommand{\ltlf}{\ensuremath{\ltl_f}\xspace}

\newcommand{\muntil}{\mathop{\U}}

\DeclareMathOperator*{\argmax}{argmax}

\begin{document}

%%
%% Rights management information.
%% CC-BY is default license.
\copyrightyear{2022}
\copyrightclause{Copyright for this paper by its authors.
  Use permitted under Creative Commons License Attribution 4.0
  International (CC BY 4.0).}

%%
%% This command is for the conference information
\conference{Joint Workshop on Statistics and Knowledge Integration for Logic, Learning, Ethical Decisions, and LLMs (SKILLED-LLMs 2026)}

%%
%% The "title" command
\title{Neuro-Symbolic Injection of LTLf Constraints in Autoregressive Reinforcement Learning Policies}

%\author{Anonymous Author}[%
%orcid=0000-0000-0000-0000,
%email=email.com
%]
%\address{Anonymous Institution}
    
%%
%% The "author" command and its associated commands are used to define
%% the authors and their affiliations.
\author{Ashkan Ansarifard}[%
orcid=0009-0005-9247-1786,
email=ansarifard.1970082@studenti.uniroma1.it
]
%\cormark[1]
%\fnmark[1]
\address{University of Rome La Sapienza, Rome, Italy}
%\address[2]{Joint Institute for Nuclear Research, 6 Joliot-Curie, Dubna, Moscow region, 141980, Russian Federation}

\author{Matteo Mancanelli}[%
orcid=0009-0004-9547-7115,
email=mancanelli@diag.uniroma1.it
]
%\fnmark[1]
%\address[3]{Vrije Universiteit Amsterdam, De Boelelaan 1105, 1081 HV Amsterdam, The Netherlands}

\author{Elena Umili}[%
orcid=0000-0002-5639-6038,
email=umili@diag.uniroma1.it
]
%\fnmark[1]
%\address[4]{University of Skövde, Högskolevägen 1, 541 28 Skövde, Sweden}

\author{Fabio Patrizi}[%
orcid=0000-0002-9116-251X,
email=patrizi@diag.uniroma1.it
]

%% Footnotes
%\cortext[1]{Corresponding author.}
%\fntext[1]{These authors contributed equally.}

%%
%% The abstract is a short summary of the work to be presented in the
%% article.
\begin{abstract}
In this work we study offline reinforcement learning (RL) under temporally extended task constraints expressed in Linear Temporal Logic over finite traces (\ltlf). Recently, transformer-based approaches such as Trajectory Transformers and Decision Transformers have been adopted to address RL as a sequence modeling problem.
%, enabling effective policy learning from fixed datasets. 
However, these methods optimize purely for reward and do not account for high-level temporal requirements.
Here, we introduce a neurosymbolic framework that injects \ltlf background knowledge into such transformer-based RL policies. 
Our approach compiles \ltlf formulas into deterministic finite automata (DFAs) and integrates them into the learning process through a differentiable representation and a logic-based loss function.
In particular, we derive differentiable satisfaction signals from DFA progression and use them as a regularization term during training. The resulting method is architecture-agnostic across different models.
We evaluate the proposed framework on navigation environments with specification suites covering combinations of safety and reachability temporal properties. Experimental results show that incorporating background knowledge not only improves constraint satisfaction, but also maintains competitive return compared to vanilla baselines.
\end{abstract}

%%
%% Keywords. The author(s) should pick words that accurately describe
%% the work being presented. Separate the keywords with commas.

%%
%% This command processes the author and affiliation and title
%% information and builds the first part of the formatted document.
\maketitle

\section{Introduction}

Safety is a central concern in reinforcement learning (RL) \cite{sutton1998reinforcement}, particularly when agents are deployed in real-world or safety-critical settings such as robotics, autonomous navigation, and decision-support systems, where undesirable behaviors may incur irreversible damage, financial loss, or human harm \cite{safe_rl_survey}. Ensuring that learned policies respect some user-defined constraints is therefore crucial for increasing the system correctness and reliability. This challenge becomes even more pronounced in offline RL \cite{offline_rl_survey}, where policies are learned exclusively from pre-collected data and cannot rely on online interaction to detect and correct unsafe behavior.

Safe decision making often requires constraints that are \emph{temporally extended}, rather than purely local or instantaneous \cite{prob_shielding_AAAI25}. Many requirements cannot be verified from a single state-action pair, but only over entire trajectories. For instance, consider the specification: \emph{"eventually reach the goal while always avoiding hazards".}
This requirement combines two temporal modalities: (i) a \emph{liveness} condition (the goal must be reached at some point before the episode ends), and (ii) a \emph{safety} condition (no hazardous state may be visited at any preceding time step). A policy that briefly steps on a hazard—even if it later reaches the goal—violates the specification. A policy that remains safe but never reaches the goal also fails. Satisfaction therefore depends on the \emph{entire trace}, not on local reward maximization alone.

In \emph{offline reinforcement learning (RL)}, enforcing such constraints is particularly challenging. Since no online interaction is allowed, constraint violations cannot be corrected via trial-and-error exploration. Sequence-model-based policies-such as Trajectory Transformers (TTs) \cite{trajectory_transformers} and Decision Transformers (DTs) \cite{decision_transformers}-are appealing in this setting because they generate trajectories autoregressively and naturally model long-range dependencies. However, they operate as probabilistic sequence generators, while task and safety requirements are typically expressed as formal logical specifications. This creates a gap between generative modeling and symbolic correctness guarantees.

In this paper, we study \emph{symbolic constraint injection} \cite{UmiliPMAI24,mezini2025neuro} for offline transformer policies using Linear Temporal Logic over finite traces (\ltlf) \cite{LTLf} as specification language. \ltlf provides a natural language for specifying unambiguous temporal requirements over episodic tasks. We compile \ltlf formulas into deterministic finite automata (DFAs) and integrate the resulting automata into the policy pipeline by training-time regularization, via differentiable soft-satisfaction signals derived from automaton progression. 
Our framework supports different transformer-based RL architectures, like TT and DT, through a shared interface that abstracts away architectural differences, and it is therefore 
%completely model-agnostic 
architecture-agnostic within the class of autoregressive sequence models. It would be possible also to extend our approach with test-time automaton-aware constrained decoding, where generation is restricted to action sequences consistent with the logical specification, but we leave that for future research.

We evaluate the proposed approach on temporally extended tasks in a navigation environment, ColourBomb \cite{prob_shielding_AAAI25}. The considered specifications span invariant safety, reachability, and combined reach-while-safe objectives. The results section is structured to systematically expose the behavior of the method: the effects of logic regularization in TT and DT, cross-benchmark trade-offs between return and logical satisfaction, and runtime overhead.
To summarize, we make the following contributions:
\begin{itemize}
    \item A unified neurosymbolic offline sequence-RL framework for injecting \ltlf constraints into transformer-based policies through automata-based modules that are architecture-agnostic.
    
    \item A canonical finite-trace evaluation stack ensuring consistent end-of-episode semantics across (i) dataset serialization, (ii) token-to-symbol mappings, and (iii) logical evaluation, thereby preventing semantic mismatches between model training and temporal-logic assessment.
    
    \item An empirical evaluation against diverse temporal specifications in navigation domains, highlighting trade-offs between performance and constraint satisfaction.
\end{itemize}
\section{Background}
\subsection{Linear Temporal Logic over finite traces}
\label{sec:ltlf}

Linear Temporal Logic (LTL) \cite{LTL} is a formal language that extends traditional propositional logic with modal operators, allowing the specification of rules that must hold \textit{through time}. Here, we use \ltl over finite traces (\ltlf) \cite{LTLf}, a popular variant of LTL which models finite, but length-unbounded, traces of executions, making it suitable for finite-horizon problems.
Given a finite set $\Sigma$ of atomic propositions, the set of \ltlf formulas $\varphi$ is inductively defined as follows:
%\begin{equation}
\[
    \varphi ::= \top \mid
    \bot
    \mid 
    a
    \mid
    \lnot\varphi
    \mid
    \varphi_1\land\varphi_2
    \mid
    X\varphi
    \mid
    \varphi_1 U\varphi_2,
\]
%\end{equation}
where $a \in \Sigma$. We use $\top$ and $\bot$ to denote $True$ and $False$, respectively. $X$ (Strong Next) and $U$ (Until) are temporal operators. Other temporal operators are $N$ (Weak Next) and $R$ (Release), defined as $N \varphi \equiv \neg X\neg\varphi$ and $\varphi_1 R \varphi_2 \equiv \neg(\neg\varphi_1 U \neg\varphi_2 )$; $G$ (globally) $G\varphi \equiv \bot R\varphi$; and $F$ (eventually) $F \varphi \equiv \top U \varphi$.

A trace $\boldsymbol{\sigma} = (\sigma_{1}, \sigma_{2}, \dots, \sigma_{T})$ is a sequence of propositional assignments to the propositions in $\Sigma$, where $\sigma_{t} \in 2^\Sigma$ is the set of all and only propositions that are true at instant $t$. Additionally, $|\boldsymbol{\sigma}| = T$ denotes the length of the trace. Since every trace is finite, $|\boldsymbol{\sigma}| < \infty$; $\epsilon$ denotes the empty trace.
Note that, if the propositional symbols in $\Sigma$ are all \textit{mutually exclusive}, i.e., the domain produces exactly one symbol true at each step, then we have $\sigma_{t} \in \Sigma$.
Let $last = |\boldsymbol{\sigma}| - 1$. We inductively define when
an \ltlf formula $\varphi$ is true at an instant $i$ (for $0 \leq i \leq last$), written $\boldsymbol{\sigma},i \models \varphi$, as follows:
\begin{itemize}
    \item $\boldsymbol{\sigma}, i \models a$ iff $a \in \sigma_i$, for $a \in \Sigma$;
    \item $\boldsymbol{\sigma}, i \models \neg \varphi$ iff $\boldsymbol{\sigma}, i \not\models \varphi$;
    \item $\boldsymbol{\sigma}, i \models \varphi_1 \land \varphi_2$ iff $\boldsymbol{\sigma}, i \models \varphi_1$ and $\boldsymbol{\sigma}, i \models \varphi_2$;
    \item $\boldsymbol{\sigma}, i \models X\varphi$ iff $i < last$ and  $\boldsymbol{\sigma}, i + 1 \models \varphi$;
    \item $\boldsymbol{\sigma}, i \models \varphi_1 \muntil \varphi_2$ iff for some $j$ such that $i \leq j \leq last$, we have $\boldsymbol{\sigma}, j \models \varphi_2$, and for all $k$ such that $i \leq k < j$, we have $\boldsymbol{\sigma}, k \models \varphi_1$.
\end{itemize}
\noindent
We say that $\boldsymbol{\sigma}$ satisfies $\varphi$, written $\boldsymbol{\sigma} \models \varphi$, if $\boldsymbol{\sigma}, 0 \models \varphi$. A formula $\varphi$ is satisfiable if it is true wrt some trace $\boldsymbol{\sigma}$, and is valid, if it is true wrt every trace $\boldsymbol{\sigma}$.

Any \ltlf formula $\varphi$ can be translated into a Deterministic Finite Automaton (DFA) \cite{LTLf} $A_\varphi = (2^\Sigma, Q, q_0, \delta, F)$, where $2^\Sigma$ is the automaton alphabet, $Q$ is the finite set of states, $q_0 \in Q$ is the initial state, $\delta: Q \times 2^\Sigma \rightarrow Q$ is the transition function, and $F \subseteq Q$ is the set of final states. 
Additionally, we recursively define the extended transition function over traces $\delta^*: Q \times (2^\Sigma)^* \rightarrow Q$ as:

\begin{equation}
%\[
\begin{array}{l}
    \delta^*(q,\epsilon) = q \\
    \delta^*(q, \sigma \cdot \boldsymbol{x}) = \delta^*(\delta(q,\sigma) , \boldsymbol{x}),
\end{array}
%\]
\end{equation}
where $\sigma \in 2^\Sigma$ is a symbol and $\boldsymbol{x} \in (2^\Sigma)^*$ is a trace.  
The automaton accepts the trace $\boldsymbol{\sigma}$ iff $\delta^*(q_0, \boldsymbol{\sigma}) \in F$, and in that case we say that $\boldsymbol{\sigma}$ belongs to the language of the automaton, denoted as $L(A_\varphi)$.  
We have that $\varphi$ and $A_\varphi$ are equivalent because, for any trace $\boldsymbol{\sigma} \in (2^\Sigma)^*$:
\begin{equation}
%\[
\boldsymbol{\sigma}\in L(A_\varphi) \iff\boldsymbol{\sigma}\vDash\varphi.
%\]
\end{equation}

\subsection{Offline Reinforcement Learning}

In Reinforcement Learning (RL), the interaction between an agent and the environment is commonly modeled as a Markov Decision Process (MDP) \cite{sutton1998reinforcement}, defined by the tuple $(S,A,t,r,\gamma)$ where $S$ is the set of states, $A$ the set of actions, $t : S \times A \times S \rightarrow [0,1]$ the transition function s.t. $\sum_{s' \in S} t(s,a,s') = 1$, $r : S \times A \rightarrow R$ the reward function, and $\gamma \in [0,1]$ the discount factor. A policy $\pi : S \rightarrow A$ maps states to actions, while the value function $V^\pi(s)$ denotes the expected discounted return obtained by following $\pi$ from state $s$. The goal of the RL agent is to learn the optimal policy $\pi^*$ maximizing expected discounted return. 

\begin{comment}
Standard MDPs assume
Markovian transitions and rewards, i.e., dependence only on
the current state. However, many real-world problems violate
this assumption [Littman et al., 2017]. In a non-Markovian
decision process, the reward function r : (S × A)∗ → R,
the transition function t : (S × A)∗ × S → [0,1], or both
may depend on the interaction history. In this work, we focus
on Non-Markovian Reward Decision Processes (NMRDPs)
[Bacchus et al., 1996], where non-Markovianity arises solely
from the reward function. Learning optimal policies in NM
RDPsis challenging, as rewards may depend on the entire se
quence of past state–action pairs, making standard RL meth
ods inapplicable. To address this issue, many approaches con
struct an augmented Markovian state representation by moni
toring the task through a labeling function L : S → P, which
maps environment states to propositional symbols. The re
sulting labeled traces are used to track task progress, typi
cally via LTL formula progression [Toro Icarte et al., 2018;
Vaezipoor et al., 2021] or equivalent automata-based repre
sentations such as Moore machines [Camacho et al., 2019;
De Giacomo et al., 2019].
\end{comment}

Traditional RL methods assume direct interaction with the environment during training. The agent collects experience by executing actions, observing state transitions and rewards, and updating its policy accordingly. This online interaction paradigm enables iterative improvement but may be costly, unsafe or even unfeasible in real-world applications. This motivates the study of offline RL \cite{offline_rl_survey}, where the agent is given a fixed dataset $\D$ of trajectories generated by some (unknown) behavior policy, and must learn a policy without additional environment interaction. In order to prevent issues with out-of-distribution data, many offline RL algorithms incorporate mechanisms to constrain learned policies to remain close to the support data distribution \cite{kumar2020conservative}.

\subsection{Reinforcement Learning as Sequence Modeling}

Some works \cite{trajectory_transformers,decision_transformers} propose an alternative perspective on RL, viewing it as a sequence modeling problem rather than a value estimation. In \cite{trajectory_transformers} a trajectory is treated as a sequence of tokens and an autoregressive model is used to learn the next token in the sequence.
Suppose we have $N$-dimensional states and $M$-dimensional actions. A trajectory $\boldsymbol{\tau}$ of horizon $T$ can be represented as
\begin{equation}
\label{eq:trace_form}
%\[
\boldsymbol{\tau} = (s_1^1, s_1^2, ..., s_1^N, a_1^1, ..., a_1^M, r_1, s_2^1, ..., r_T)
%\]
\end{equation}
\noindent
Subscripts on all tokens denote timestep and superscripts on states and actions denote dimension (i.e., $s_t^i$ is the $i^\text{th}$ dimension of the state at time $t$). An autoregressive model is capable of estimating the probability of the $t^\text{th}$ token $x_{t}$ given the set of all past tokens $\boldsymbol{x_{<t}}$:
\begin{equation}
\label{eq:next_act_pred}
P(x_{t} \mid x_{1}, x_{2}, \dots, x_{t-1}) = P(x_{t} \mid \boldsymbol{x_{<t}}).
\end{equation}
\noindent
Instead of learning value functions or policies, one can learn the joint distribution
\begin{equation} \label{eq:seq_joint_prob}
P(\boldsymbol{x}) = \prod_{i=1}^{|\tau|} P(x_{i} \mid \boldsymbol{x_{<i}}).
\end{equation}

At each generation step, a token must be \textit{sampled} from the next activity probability. A common way of selecting the next token to feed into the autoregressor is to \textit{greedily} choose it by maximizing the next token probability at each step, as follows:
%\[
\begin{equation}
{x}_{t} = \argmax_x P(x_{t} \mid \boldsymbol{x}_{<t}), \quad \forall t < |\tau|.
\end{equation}
%\]
\noindent
In general, this greedy search strategy is sub-optimal, because it may not produce the trace that maximizes the joint probability distribution in Equation \ref{eq:seq_joint_prob}. Other sub-optimal sampling strategies commonly used for this task include Beam Search, Random Sampling, and Temperature Sampling~\cite{ramamaneiro24}.

Trajectory Transformer (TT) models are transformer-based autoregressive architectures well suited for addressing RL as sequence modeling. In the offline setting, such models can be trained via maximum likelihood using teacher forcing, by maximizing the log-probability of trajectories in the dataset. At inference time, planning can be performed by sampling or beam search, optionally biasing candidate trajectories according to cumulative reward or reward-to-go estimates. This paradigm unifies dynamics modeling, policy learning, and behavior regularization within a single sequence model.
Note that this approach focuses exclusively on maximizing return and modeling the data distribution, but they may violate the constraints that emerge from (or we want to impose to) the domain. This motivates our neurosymbolic framework guiding the learned trajectory distribution toward constraint satisfaction.
\section{Related Work}
 
\subsection{Offline RL as sequence modeling}

Offline Reinforcement Learning \cite{offline_rl_survey} is frequently addressed by techniques that are based on value functions \cite{kumar2020conservative,kostrikov2021offline}, goal-conditioned behavior cloning \cite{nair2018visual,ding2019goal} and reward-conditioned behavior cloning \cite{kumar2019reward,srivastava2019training}.
As previously explained, recent works show how this problem can be tackled as a sequence model problem. In \cite{emmons2021rvs} RL via supervised learning is analyzed, this work shows that sometimes pure supervised learning is competitive wrt more complex architectures, given careful tuning of the policy capacity. In contrast, other recent works have adopted the use of Transformers \cite{transformers} as sequence prediction models, and they have provided a promising direction for offline RL, which is also flexible enough to be combined with mentioned techniques. 

In particular, Trajectory Transformers (TTs) \cite{trajectory_transformers} and Decision Transformers (DTs) \cite{decision_transformers} follow this idea by using beam-search planning and reward conditioning, respectively. Notably, these architectures are further studied combining techniques based on Dynamic Programming (such as Q-learning) \cite{yamagata2023q}, online fine-tuning \cite{zheng2022online}, few-shot policy generalization \cite{xu2022prompting}, and automatically-generated waypoints \cite{badrinath2023waypoint}. There are also works that do not build directly on TTs or DTs, such as \cite{huang2024diffusion}, where Transformers are used to overcome inefficiencies of Diffusion Models.

\subsection{Safe Reinforcement Learning}

Safe RL focuses on methods for ensuring that learned policies satisfy some given safety constraints. Online RL techniques \cite{achiam2017constrained,tessler2018reward} typically model safety requirements as constraints on the expected cost, and enforce safety through constrained optimization or Lagrangian formulations during policy learning. Recently, also offline Safe RL is gaining popularity, including techniques for both soft constraints \cite{xu2022constraints} and hard constraints \cite{zheng2024safe}. Adaptations of DTs and TTs are also studied in the context of offline Safe RL \cite{liu2023constrained,wang2024safe,guo2024temporal,zhang2023saformer}, confirming the growing interest in these models in recent years. These works share a common formulation grounded in Constrained MDPs, where safety is enforced by conditioning the policy on cost signals and constraint thresholds. Despite their differences, all these approaches operate by modifying the conditioning signals or relabeling trajectory returns to steer generation toward safe behavior. In contrast, our method leaves the trajectory data and its reward/cost structure untouched and instead regularizes the training objective with a differentiable logic loss derived from DFA progression, injecting temporal-logic knowledge directly into the learning process rather than into the data representation.
Recent work \cite{tian2023reinforcement} also addresses RL using TTs and \ltlf constraints, but their approach differs from ours as it relies on reward shaping derived from DFA states to imitate trajectories.
%They relies on reward shaping derived from DFA states and trains the model to imitate trajectories generated with these shaped rewards. In contrast, our method integrates temporal logic into the training objective through logic-based regularization, enabling constraint-aware learning without modifying the reward function.

An alternative line of work enforces safety through shielding \cite{alshiekh2018safe,jansen2018safe,belardinelli2025probabilistic}, where a runtime monitor prevents the agent from executing actions that may lead to unsafe states. 
While shielding provides strong safety guarantees, it typically assumes the availability of an explicit environment model, it is primarily designed for on-policy learning, and it is structurally limited to the "safety fragment" of formal specifications. In contrast, our method is well suited for offline RL, does only require knowledge of the constraints without assuming any other knowledge of the environment, and can be used to impose any LTLf formula as a constraint, also the ones outside the safety fragment. 

\subsection{Temporally-constrained sequence generation}
\begin{comment}
% old section on RL and LTLf removed

Temporal logic formalisms, such as \ltl \cite{LTL} and \ltlf \cite{LTLf}, are often used in RL with non-Markovian rewards to define temporally extended goals and constraints. These allow to recover the Markovian property by conditioning the policy on the state of an automaton constructed from the formula. Well-known examples include Restraining Bolts \cite{de2019foundations,de2020restraining} and Reward Machines \cite{icarte2022reward,camacho2019ltl,toro2018teaching}. \ltl-based methods are also employed for multi-task RL \cite{vaezipoor2021ltl2action,jackermeier2024deepltl}, reward-shaping techniques \cite{hasanbeig2019certified,shao2023sample}, task inference \cite{hasanbeig2024symbolic}, and shielding \cite{varricchione2024pure,corsi2024verification}.
\end{comment}

Recently, there has been increasing interest in constraining autoregressive sequence generation through logical knowledge. Applications span Cyber-Physical Systems~\cite{STLnet}, Business Process Management~\cite{UmiliPMAI24,DiFrancescomarino17}, and natural language generation with Large Language Models (LLMs)~\cite{trident,llm_beam_search_1,llm_beam_search_2,llm_sampling1,montecarlo_llm,pseudosemantic_loss}.
Most prior work incorporates constraints at \emph{test time}, guiding suffix generation via constrained beam search~\cite{DiFrancescomarino17,trident,llm_beam_search_1,llm_beam_search_2}, auxiliary tractable models~\cite{llm_aux_mod_1,llm_aux_mod_2}, or conditioned sampling~\cite{llm_sampling1,montecarlo_llm}.
In contrast, we enforce constraints at \emph{training time} through an auxiliary logical loss. Only few approaches follow this direction~\cite{STLnet,UmiliPMAI24,pseudosemantic_loss}. Among them,~\cite{UmiliPMAI24,pseudosemantic_loss} estimate the probability of satisfying a formula via pseudoprobabilities or Monte Carlo, while STLnet~\cite{STLnet} relies on a student–teacher scheme that is difficult to apply in discrete domains~\cite{UmiliPMAI24}.

A key limitation of existing methods is that they target \emph{homogeneous} sequences (e.g., text), assuming that logical formulas are directly expressed over the generated tokens. While reasonable for language, this assumption is limiting in RL settings.
Here, we consider trajectories of states, actions, and rewards (as formalized in Eq.~\ref{eq:trace_form}), where constraints are naturally specified over abstract, high-level \emph{symbols} that must be grounded in subsequences representing states or state–action pairs.
Our approach builds on~\cite{UmiliPMAI24,mezini2025neuro}, which approximate \ltlf satisfaction via Monte Carlo and use it as a training loss. We extend this framework to Decision and Trajectory Transformers for safe offline RL applications.

\section{Method}
\subsection{Problem Formulation}
In offline RL, the agent is provided with a fixed dataset $\mathcal{D}=\{\boldsymbol{\tau}^{(i)}\}_{i=1}^N$ of previously collected trajectories.
%Let $\mathcal{D}=\{\boldsymbol{\tau}^i\}_{i=1}^N$ be an dataset of trajectories. Each trajectory is a finite sequence of transitions and terminal marker, i.e. it has the form of (\ref{eq:trace_form}) plus a special symbol $\texttt{END}$ at the end of the trace. At each timestamp, together with state dimensions, action dimensions and rewards, we can have an optional environment-specific auxiliary fields (e.g., safety cost).
%Let $\mathcal{D}=\{\boldsymbol{\tau}^{(i)}\}_{i=1}^N$ be a dataset of trajectories. 
Each trajectory is a finite sequence of transitions and an \emph{end-of-trace} symbol $\texttt{EOT}$, i.e. $\boldsymbol{\tau} = (y^1,\ldots,y^T,\texttt{EOT})$, where each transition token $y^j$ group encodes state, action, reward, and optional environment-specific auxiliary fields (e.g., safety cost). 
%In addition to the return signal, 
Here, we also consider some prior knowledge about the task, expressed as \ltlf constraints, to capture high-level behavioral requirements. Given an \ltlf formula $\varphi$, our objective is to learn a trajectory generation policy that maximizes both the expected return of the policy and the probability that generated trajectories satisfy $\varphi$.
We define atomic propositions through an environment-specific extractor 
\[
\Pi(y^t)\in\{0,1\}^K
\]
\noindent
and the \ltlf formula $\varphi$ is defined over these $K$ propositional symbols. Note that, in our experiments, proposition extraction is \textit{state-centric}, i.e. $\varphi$ is defined only over state dimensions, but our approach is fully general and supports also imposing constraints on action, state-action pairs and auxiliary-terms.

A logic adapter is used to define the authoritative token schema and the unique end token identifier shared by dataset builders, token-to-symbol mapping, DFA rollout, and logic-loss evaluation. 
Satisfaction evaluation follows the standard \ltlf semantics (see Sec. \ref{sec:ltlf}) and is therefore defined on complete finite traces; traces without an end marker are treated as incomplete and unsatisfied. This design avoids inconsistent termination handling across modules and ensures agreement between hard DFA evaluation and differentiable soft evaluation.

As explained in the background, we assume an autoregressive neural model $f_\theta$ with trainable parameters $\theta$ that estimates an approximation $P_\theta$ of the probability of the next event $\tau_t$ given a trace of previous events $\boldsymbol{\tau}_{<t}$ (Equation~\ref{eq:next_act_pred}):
\begin{equation} 
\label{eq:RNN}
%\[
\begin{array}{ll}
     \tilde{y}_t = f_\theta(\boldsymbol{\tau}_{<t}) \\
     P(\tau_{t} = v_i \mid \boldsymbol{\tau}_{<t}) \approx \tilde{y}_t[i].
\end{array}
%\]
\end{equation}
where, $\tau_{t}$ denotes the token at position $t$, and $v_i$ is an element of the token vocabulary. Note that we do not make any assumptions about the neural model, except that it can estimate the probability of the next activity given a sequence of previous ones. As a result, our approach is entirely \textit{model-agnostic} and can be readily applied to any autoregressive model (provided that the logic adapter is changed consistently). The model parameters are typically trained using a supervised loss $L_{\mathcal{D}}$, evaluated on a dataset $\mathcal{D}$ of ground-truth traces obtained by observing the process. The loss for a trace $\boldsymbol{\tau} \in \mathcal{D}$ of length $T$ is defined as follows:
\begin{equation} \label{eq:sup_loss}
    L_{\mathcal{D}}(\boldsymbol{\tau}) = \frac{1}{T} \sum_{t=1}^T \text{cross-entropy}(f_\theta(\boldsymbol{\tau}_{<t}), \tau_{t}) .
\end{equation}
%This loss trains the network to predict the next symbol in a trace so as to closely mimic the data in the dataset.

Here, the goal is 
%Our goal is for the language generated by the autoregressor $f_\theta$ to be strictly contained within the language of strings accepted by the \ltlf formula $\varphi$, denoted as $\calL(A_{\varphi})$. However, the language produced by the network is \emph{unbounded}, as it is only \emph{softly assigned}: in other words, the network can generate any possible string, each with a different probability.
%
%Our method therefore aims 
to maximize the probability $P_{\theta \vDash \varphi}$ that traces $\boldsymbol{\tau} \sim P_\theta$, sampled from the autoregressor, satisfy the specification:
\begin{equation} \label{eq:logic_prob}
P_{\theta \vDash \varphi} = \mathbb{E}_{\boldsymbol{\tau} \sim P_\theta}[\boldsymbol{\tau} \vDash \varphi] = 
\sum_{\boldsymbol{\tau}} P_\theta(\boldsymbol{\tau}) \, \mathbb{I}\{\boldsymbol{\tau} \vDash \varphi\}.
\end{equation}

\noindent
Computing this probability exactly is infeasible, since it would require to enumerate all possible traces of maximum length $T$. In \cite{UmiliPMAI24,mezini2025neuro} a differentiable procedure to approximate the previous probability at each optimization step of the autoregressor is designed. In particular, they introduce a logic loss function $L_\varphi$, and compute the training objective as a linear combination of the logic and the supervised loss:
\begin{equation}
\label{eq:loss_balance}
    L = \alpha L_\varphi + (1 - \alpha) L_{\mathcal{D}},
\end{equation}
with $\alpha$ being a trade-off parameter between 0 and 1 that balances the influence of each loss on the training process.
Here we adapt the logic loss computation to offline RL domains and apply it at training time, together with other techniques at test time to maximize adherence with the constraints.

\subsection{Automata-Based Satisfaction and Logic Loss}

Now, we briefly discuss how to approximate the target probability in Equation \ref{eq:logic_prob}. Specifically, \cite{mezini2025neuro} uses a Monte Carlo estimation by sampling a set of complete traces $\{\boldsymbol{\tau}^{(1)}, \boldsymbol{\tau}^{(2)}, \ldots, \boldsymbol{\tau}^{(N)}\} \sim P_\theta$ according to the distribution learned by the autoregressor, and compute an approximation of the target probability as follows:
\begin{equation} \label{eq:montecarlo_approx}
    \hat{P}_{\theta \vDash \varphi} = \frac{1}{N} \sum_{i=1}^N \mathbb{I} \{ \boldsymbol{\tau}^{(i)} \vDash \varphi \}.
\end{equation}
Note that the sampled traces $\boldsymbol{\tau}^{(i)}$ are generated by a NN that returns a probability distribution at each time step, and thus they are replaced by a probabilistic counterpart $\tilde{\boldsymbol{\tau}}^{(i)}$, where symbols are sampled from that probability distribution. Moreover, the indicator function $\mathbb{I}$ in Equation~\ref{eq:montecarlo_approx} is replaced by a method that computes the (probabilistic) compliance of a probabilistic trace with the knowledge. To achieve this, we can leverage on \emph{(i)} the Gumbel-Softmax sampling, that generates differentiable, near one-hot suffixes during training and \emph{(ii)} DeepDFA \cite{deepdfa_ecai2024}, that encodes temporal logic properties as a recurrent layer, enabling efficient and differentiable evaluation of logical constraints.

\paragraph{Gumbel-Softmax sampling.} To sample from the probability distribution returned by a NN, but still maintain differentiability, we can use the Gumbel-Softmax sampling, which produces approximations of discrete symbols as one-hot-like vectors. Given the predicted token distribution $\tilde{y}_t$, we get 
\begin{equation}
\tilde{x}_t = \text{softmax} \left( \frac{\log (\tilde{y}_t) +G}{temp} \right)
\end{equation}
\noindent
where $G$ is a Gumbel noise vector and $temp$ is a temperature parameter controlling the sharpness of the distribution. These soft tokens are then fed into the differentiable DFA, which computes the probability that the sampled trajectory satisfies the LTLf specification. As $temp \rightarrow 0$, the output approaches a discrete one-hot vector, while for $temp = 1$, it remains close to the original continuous probabilities in $\tilde{y}_t$. Since the next activity is only \emph{probabilistically} grounded, we denote it as $\tilde{\tau}_t$.

\paragraph{DeepDFA.}
As explained in Sec. \ref{sec:ltlf}, each \ltlf constraint $\varphi$ can be compiled into an equivalent DFA $A_\varphi$. We extend the DFA to handle the special $\texttt{EOT}$ symbol, such that only traces including $\texttt{EOT}$ are accepted. There exists automatic translation tools, such as \texttt{ltlf2DFA}~\cite{fuggitti-ltlf2dfa}, that can be used off-the-shelf.
Many alternatives exist in the literature to evaluate continuous satisfaction of temporal constraints over sequences of \textit{soft} (probabilistic or fuzzy) symbol assignments \cite{umili_kr23,deepdfa_ecai2024,DonadelloFIMM25,NesyA}. In our work we used DeepDFA \cite{deepdfa_ecai2024}, following what has been done in \cite{UmiliPMAI24,mezini2025neuro}. DeepDFA is a neural, probabilistic relaxation of a standard DFA, where the automaton is represented in matrix form and the input symbols, states, and outputs are \textit{probabilistically grounded}. This allows us to compute the probabilistic compliance $P_{\mathrm{DDFA}}(\tilde{\boldsymbol{\tau}}^{(i)} \vDash \varphi)$ of a sampled trace $\tilde{\boldsymbol{\tau}}^{(i)}$ with the knowledge $\varphi$. Therefore, Equation~\ref{eq:montecarlo_approx} becomes:
    \begin{equation} \label{eq:montecarlo_approx_ddfa}
    \hat{P}_{\theta \vDash \varphi} = \frac{1}{N} \sum_{i=1}^N P_{\mathrm{DDFA}}(\tilde{\boldsymbol{\tau}}^{(i)} \vDash \varphi).
\end{equation}
\noindent
Finally, we are able to compute the logic loss $L_\varphi$, which enforces the satisfaction of prior knowledge over entire traces:
\begin{equation}
L_\varphi = - \log \left( \hat{P}_{\theta \vDash \varphi} \right).
\end{equation}
%and we use it in Equation \ref{eq:loss_balance}. 

\begin{comment}
\subsection{TT: Logic-Regularized Training and Automaton-Aware Decoding}
TT models trajectories autoregressively. We use standard likelihood training and optionally add a logic regularization term weighted by $\alpha$ (swept in experiments), using soft satisfaction computed from sampled/generated traces. We also support multiple DFA composition modes (\texttt{single}, \texttt{product}, \texttt{multi}) for composite specifications.

At evaluation, we compare:
\begin{itemize}
\item greedy decoding,
\item unconstrained beam search,
\item automaton-aware constrained beam search.
\end{itemize}
Each beam maintains a DFA state. We support soft satisfaction reranking and hard pruning of beams that enter identified rejecting sink states.

\subsection{DT: Baseline and Logic-Enhanced Variants}
We consider the following DT family methods:
\begin{itemize}
\item \textbf{DT vanilla:} greedy inference baseline.
\item \textbf{DT-LTF-1:} training-time logic regularization with rollout-based soft satisfaction (logic weight \texttt{logic\_alpha} and rollout controls).
\item \textbf{DT-LTF-2:} constrained test-time action selection (\texttt{dt\_mode constrained}) using candidate actions, short-horizon lookahead, DFA state updates, and optional hard pruning.
\item \textbf{DT-LTF-3:} offline kNN suffix continuation planner (\texttt{dt\_mode knn}) scoring retrieved continuations with return and satisfaction proxies.
\end{itemize}
For DT evaluation, we additionally report a random policy baseline as a reference.
\end{comment}

\section{Experimental Evaluation}

To evaluate the proposed framework, we conduct experiments\footnote{The code is publicly available at: https://github.com/ashkanans/nesy\_rl\_test.} on a controlled navigation benchmark designed to simulate safety-critical decision making under temporally extended constraints. The experiments focus on understanding how the proposed logic regularization influences the behavior of transformer-based policies and how different levels of logical guidance affect the trade-off between task performance and constraint satisfaction.

\subsection*{Environment}

\begin{figure}[!th]
\centering
\includegraphics[width=0.68\linewidth]{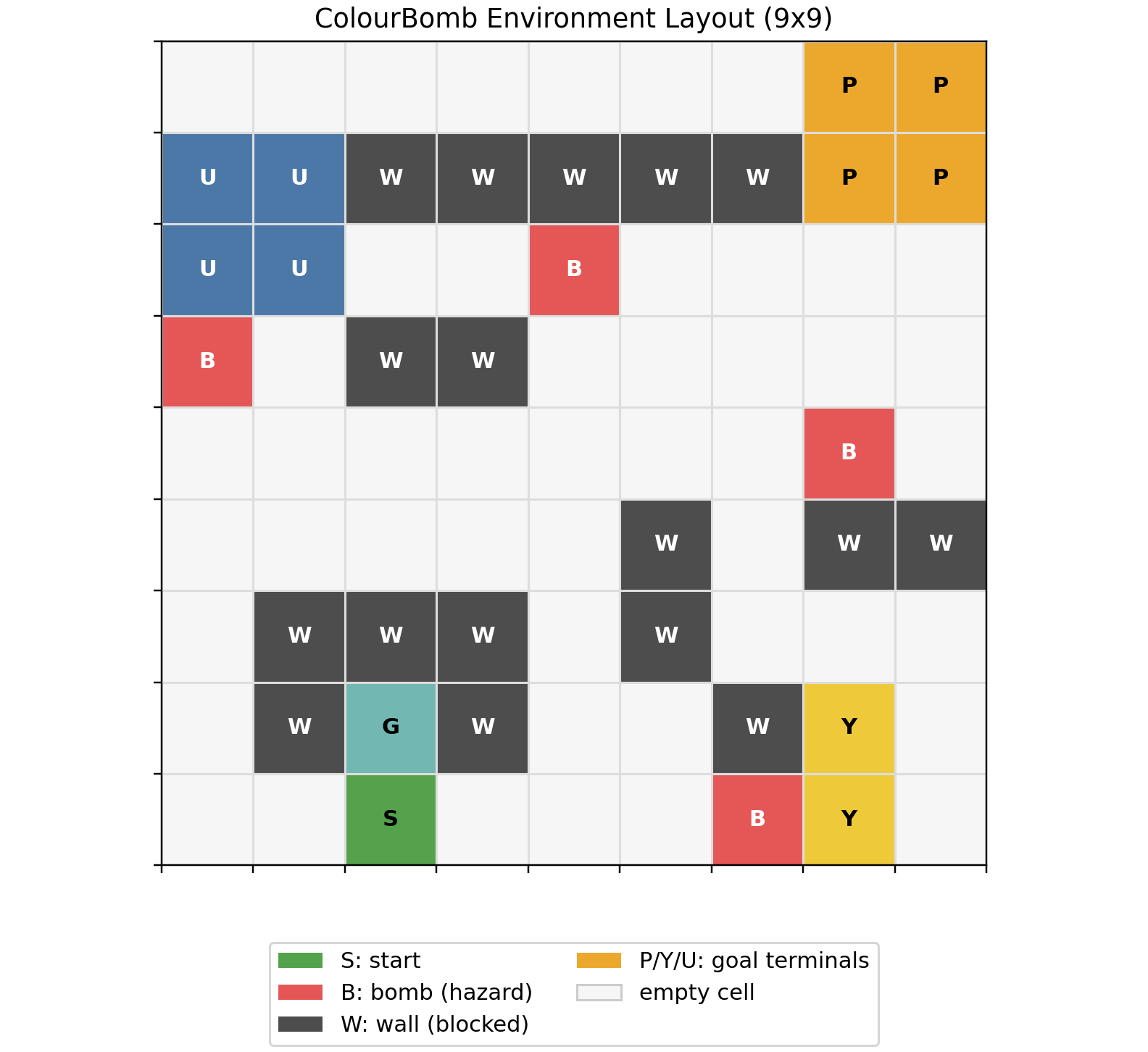}
\caption{ColourBomb environment structure.}
\label{fig:cb_env_layout}
\end{figure}

We evaluate our approach on the ColourBomb environment \cite{prob_shielding_AAAI25}, a grid-based navigation domain illustrated in Figure \ref{fig:cb_env_layout}.
The environment contains:
\begin{itemize}
    \item \textbf{Start cell (S): } the initial location of the agent.
    \item \textbf{Goal cells (P/Y/U): } terminal states (identified by different colours) that provide positive reward and end the episode.
    \item \textbf{Bomb cells (B): } hazardous states that terminate the episode with negative reward.
    \item \textbf{Walls (W): } non-traversable cells that block movement.
\end{itemize}

The agent must navigate from the start location toward one of the goal cells while avoiding bombs and obstacles. The environment is episodic with a finite horizon. Each step incurs a small penalty to encourage efficient navigation, while reaching a goal yields positive reward and entering a bomb cell yields negative reward. This environment is particularly suitable for testing our framework, and its flexibility enables the construction of different \ltlf constraints with increasing levels of complexity.

\subsection*{Temporal Specifications}

We evaluate the system using two representative families of \ltlf specifications:

\begin{itemize}
    \item \textbf{Safety:} requires the agent to avoid bomb cells throughout the entire trajectory
\[
G(\neg bomb)
\]
    \item \textbf{Reach-while-Safe:} requires the agent to eventually reach a goal while always avoiding bomb
\[
G(\neg bomb) \land F(goal)
\]
\end{itemize}

\begin{figure}[!th]
\centering
\includegraphics[width=0.34\linewidth]{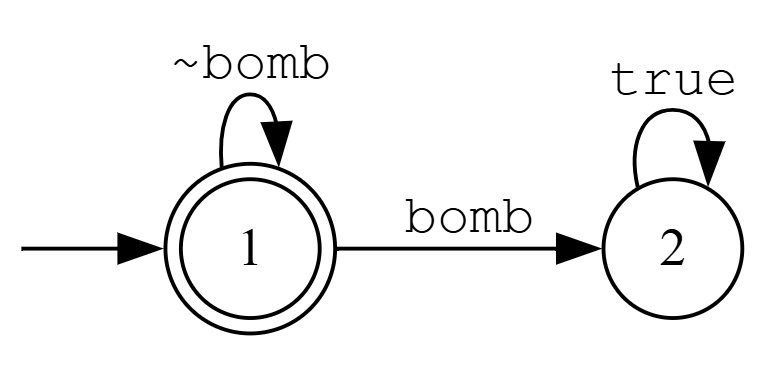}
\includegraphics[width=0.65\linewidth]{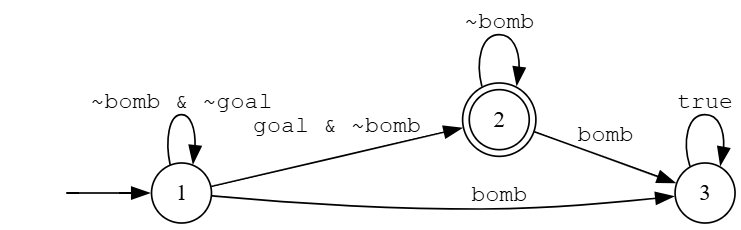}
\caption{DFAs for Safety and Reach-while-Safe constraints}
\label{fig:dfas}
\end{figure}

\noindent
Note that $F(goal)$ is a liveness property in temporal logics terminology. Note also that, since we assume \emph{(i)} goal is terminal, \emph{(ii)} goal is not a bomb, and \emph{(iii)} traces stop when goal is reached, then we could rewrite the constraint as $(\neg bomb) \,U\, goal$,
which ensures that the agent remains in safe states until a goal is reached. We show the automata for to these two specifications in Fig. \ref{fig:dfas}.

\noindent
These specifications allow us to analyze the behavior of the method under both purely safety-oriented and mixed safety-and-performance objectives. One could also consider more complex formulas, for instance requiring the agent to visit coloured states in a given order.
Notably, our framework goes beyond most papers on Safe RL, allowing the specification of arbitrary temporal constraints. In future work, we aim to explore the use of combined formulas drawn from \textit{Declare} patterns~\cite{declare2006}.

\subsection*{Methods Compared}

We evaluate variants of the Trajectory Transformer and Decision Transformer architectures with different levels of logical regularization. We consider vanilla TT and DT as the baseline model, without any logical regularization. This corresponds to setting $\alpha=0$ to Equation \ref{eq:loss_balance}. Then, we impose logic regularization, that is, the training objective combines the standard sequence-prediction loss with the proposed logic loss computed by the use of differentiable DFAs. In order to control the weight of logic regularization and study how it affects return and constraint satisfaction, we vary the trade-off parameter $\alpha \in \{0.01,0.05,0.1,0.2,0.4, 0.6, 0.8\}$. All models use greedy decoding at inference time to isolate the effect of the logic regularization during training. 

\subsection*{Evaluation Metrics}

We evaluate policies using several metrics capturing both task performance and logical compliance:
\begin{itemize}
    \item \textbf{Return: } the average episodic reward obtained by the agent.
    \item \textbf{Satisfaction Rate: } the fraction of trajectories that satisfy the \ltlf specification when evaluated using the compiled DFA.
    \item \textbf{Goal Rate: } the proportion of episodes in which the agent successfully reaches a goal state.
    \item \textbf{Bomb Hit Rate: } the fraction of trajectories that terminate by entering a bomb cell.
\end{itemize}

\subsection*{Experiments on Decision Transformers}
Here we report results obtained by adding the logic loss to Decision Transformers. The goal of these experiments is to assess the effectiveness of logic regularization on both safety and task performance.
\paragraph{Invariant Safety.}
We report Decision Transformer (DT) alpha-sweep results with greedy decoding.
For the invariant safety specification $G(\neg bomb)$, low-to-mid regularization strengths ($\alpha \leq 0.4$) do not improve behavior over vanilla DT: the agent still hits bombs and never reaches the goal.
At higher regularization, both $\alpha=0.6$ and $\alpha=0.8$ achieve full safety satisfaction and zero bomb-hit rate.
However, $\alpha=0.6$ is overly conservative (goal rate $=0$), while $\alpha=0.8$ preserves safety and also reaches the goal, yielding the strongest return.

\begin{figure}[!th]
\centering
\includegraphics[width=\linewidth]{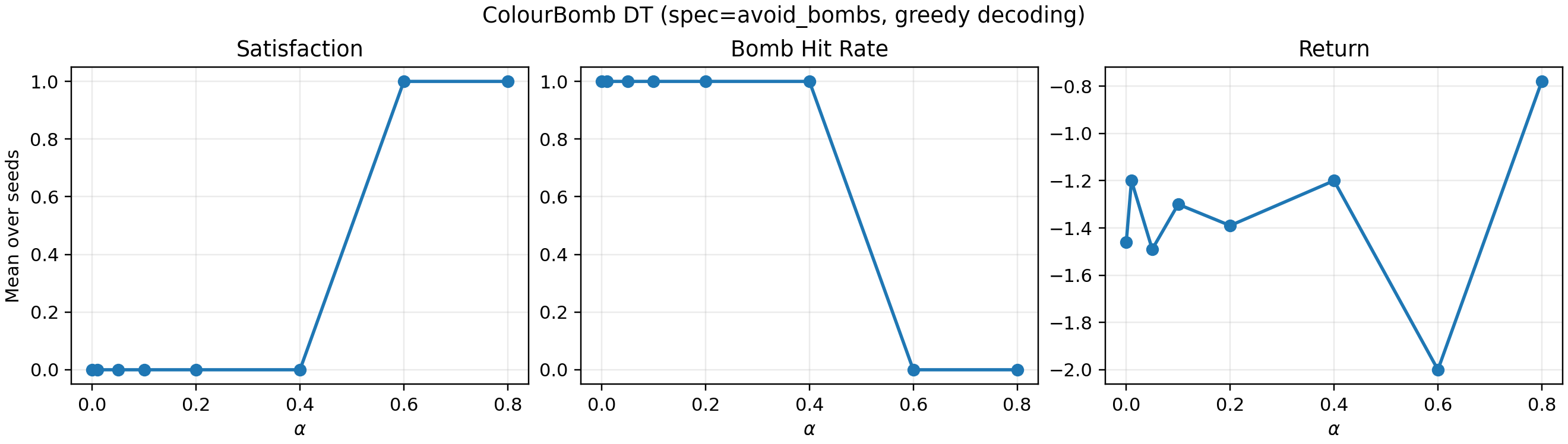}
\caption{DT results for the invariant safety constraint as a function of $\alpha$. Plots report the mean and standard deviation over three runs.}
\label{fig:cb_dt_alpha_avoid}
\end{figure}

\paragraph{Reach-while-Safe.}
For the conjunctive specification $G(\neg bomb) \land F(goal)$, the setting is harder because safety and goal achievement must hold together.
As in the invariant case, $\alpha \leq 0.4$ does not change the baseline failure mode.
At $\alpha=0.6$, DT removes bomb hits but still fails to satisfy the full formula due to zero goal-reaching.
Only $\alpha=0.8$ jointly achieves safety and goal attainment, giving the best satisfaction and return. These results are consistent with the findings in the invariant security case.

\begin{figure}[!th]
\centering
\includegraphics[width=0.9\linewidth]{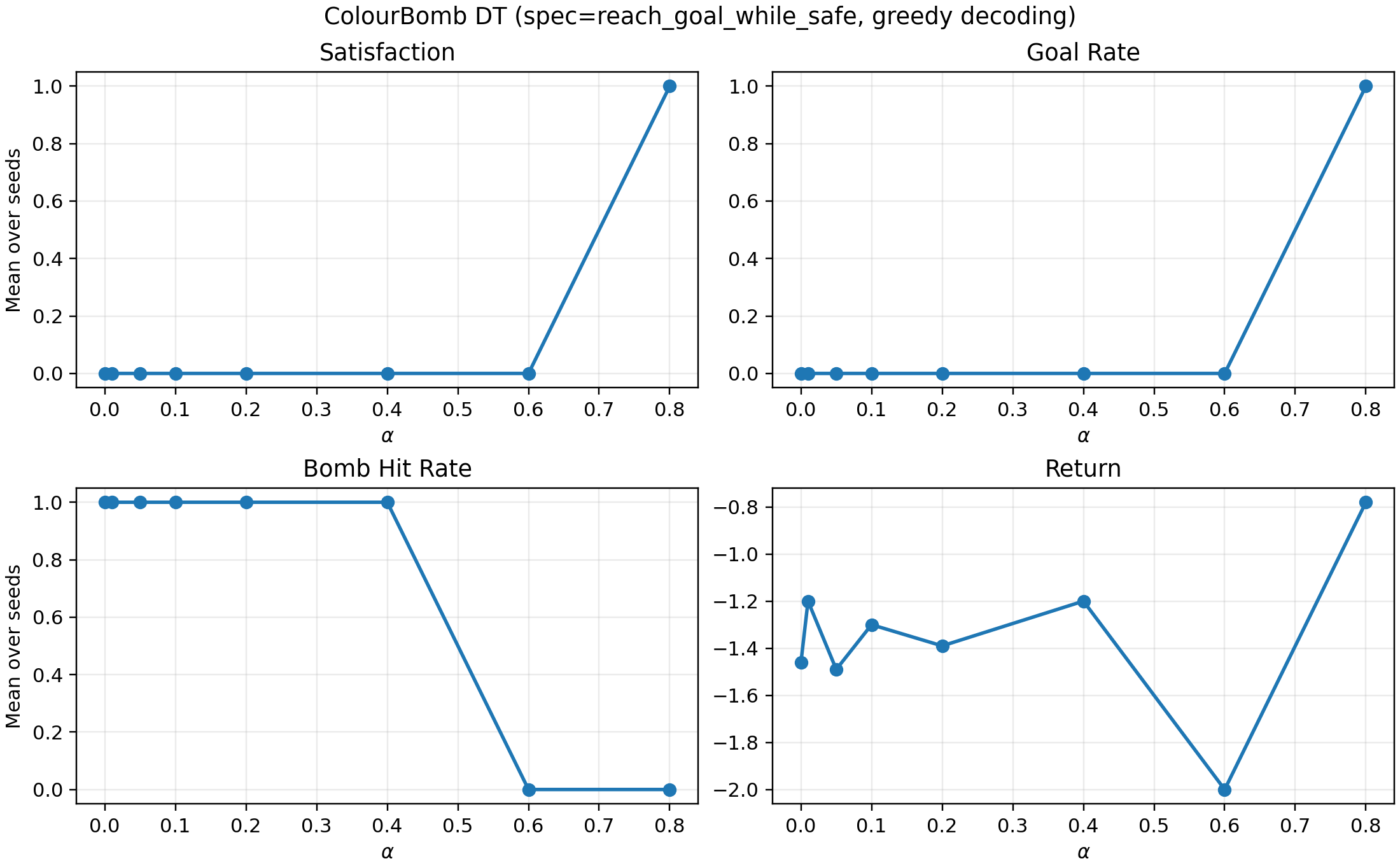}
\caption{DT results for the Reach-while-Safe constraint as a function of $\alpha$. Plots report the mean and standard deviation over three runs.}
\label{fig:cb_dt_alpha_tradeoff}
\end{figure}

\begin{comment}
\paragraph{Cross-Specification Comparison.}
Figure \ref{fig:cb_dt_cross_spec} compares the best DT configuration for each specification.
In this sweep, both objectives select $\alpha=0.8$ as the strongest operating point, with high satisfaction, zero bomb-hit rate, and successful goal completion.
Figure \ref{fig:cb_dt_metrics_bar_example} reports the full per-metric bar summary across all $\alpha$ values for both specifications.

\begin{figure}[!th]
\centering
\includegraphics[width=0.85\linewidth]{figures/colourbomb/cb_dt_spec_comparison_goal_vs_safety.png}
\caption{DT best-performing value of $\alpha$ for each specification. Results are aggregated over three runs.}
\label{fig:cb_dt_cross_spec}
\end{figure}

\begin{figure}[!th]
\centering
\includegraphics[width=\linewidth]{figures/colourbomb/cb_dt_metrics_bar_example.png}
\caption{DT summary of performance metrics for both specifications, comparing vanilla and logic-regularized variants.}
\label{fig:cb_dt_metrics_bar_example}
\end{figure}
\end{comment}

\paragraph{Quantitative Summary.}
Table~\ref{tab:cb_dt_key_results} highlights a sharp transition in DT behavior under logic regularization.
While vanilla DT completely fails—yielding zero satisfaction and always hitting bomb states—high logical regularization ($\alpha=0.8$) improves significantly the behavior of the model. In this setting, the policy simultaneously achieves perfect satisfaction, zero bomb-hit rate, and consistent goal reaching across both specifications, leading to a substantial improvement in return. 

\begin{table}[!th]
\centering
\begin{tabular}{l l c c c c}
\toprule
Spec & Setting & Satisfaction & Goal & Bomb Hit & Return \\
\midrule
\texttt{safety} & vanilla ($\alpha=0$) & 0.000 & 0.000 & 1.000 & -1.460 \\
\texttt{safety} & logic ($\alpha=0.8$) & \textbf{1.000} & \textbf{1.000} & \textbf{0.000} & \textbf{-0.780} \\
\midrule
\texttt{reach-while-safe} & vanilla ($\alpha=0$) & 0.000 & 0.000 & 1.000 & -1.460 \\
\texttt{reach-while-safe} & logic ($\alpha=0.8$) & \textbf{1.000} & \textbf{1.000} & \textbf{0.000} & \textbf{-0.780} \\
\bottomrule
\end{tabular}
\caption{Decision Transformer key results.}
\label{tab:cb_dt_key_results}
\end{table}

\subsection*{Experiments on Trajectory Transformers}
\begin{figure}[!th]
\centering
\includegraphics[width=0.9\linewidth]{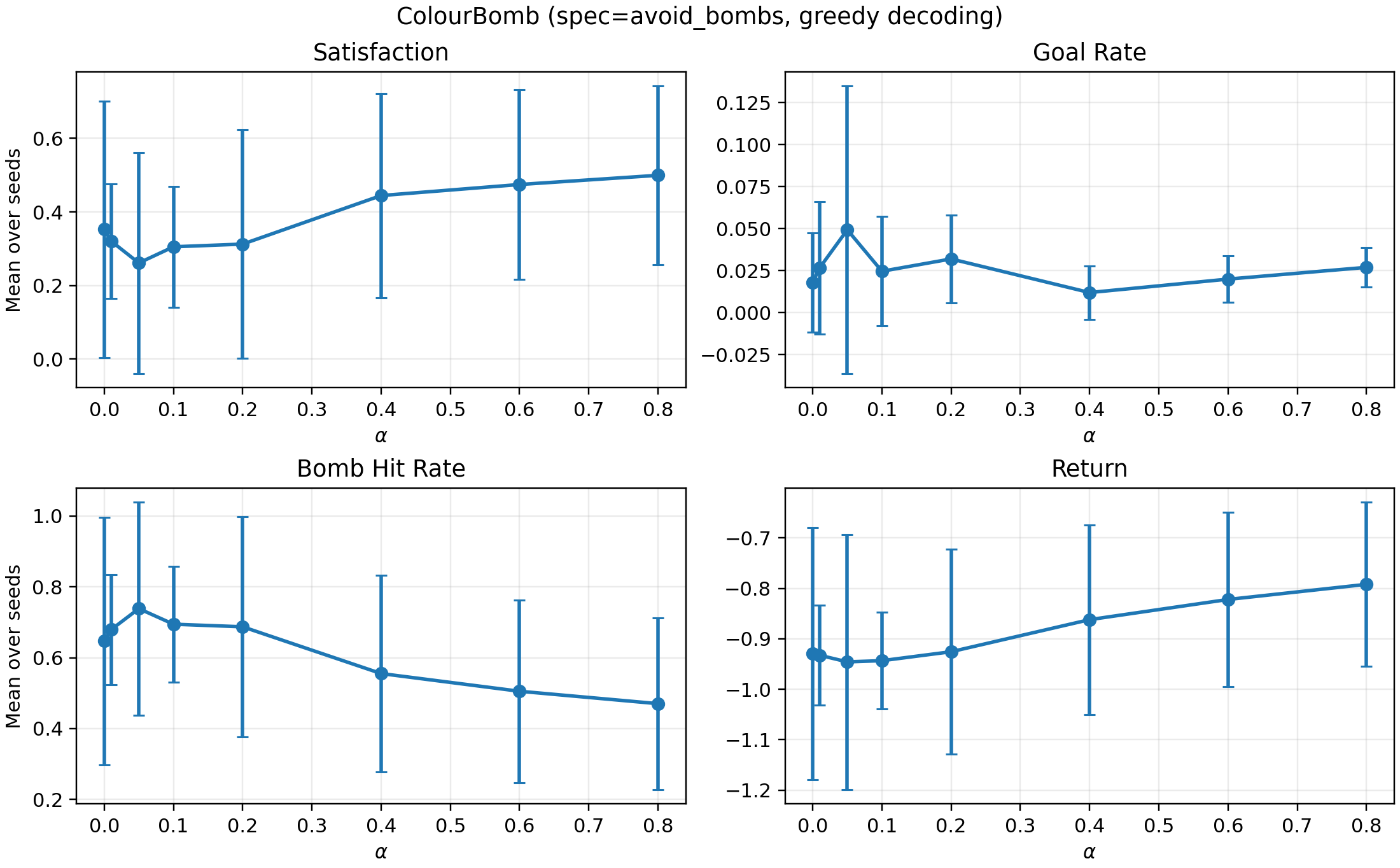}
\caption{Results under the safety constraint as a function of the weighting loss parameter $\alpha$. Plots report the mean and standard deviation over three runs.}
\label{fig:cb_alpha_avoid}
\end{figure}

We now present preliminary results obtained by our implementation based on Trajectory Transformers. 
\paragraph{Invariant Safety.}
We first analyze the invariant safety specification $G(\neg bomb)$. Figure \ref{fig:cb_alpha_avoid} shows that small values of $\alpha$ do not improve safety compared to the baseline, while larger values of the logic regularization provide measurable benefits. Setting $\alpha=0.8$ achieves the highest satisfaction rate. In general, the logic-regularized policy reduces the probability of entering bomb cells and improves overall compliance with the specification. At the same time, it does not affect task performance, for which our method maintains comparable results. Notably, setting $\alpha=0.8$ improves also the return with respect to the baseline.

\begin{figure}[!th]
\centering
\includegraphics[width=0.9\linewidth]{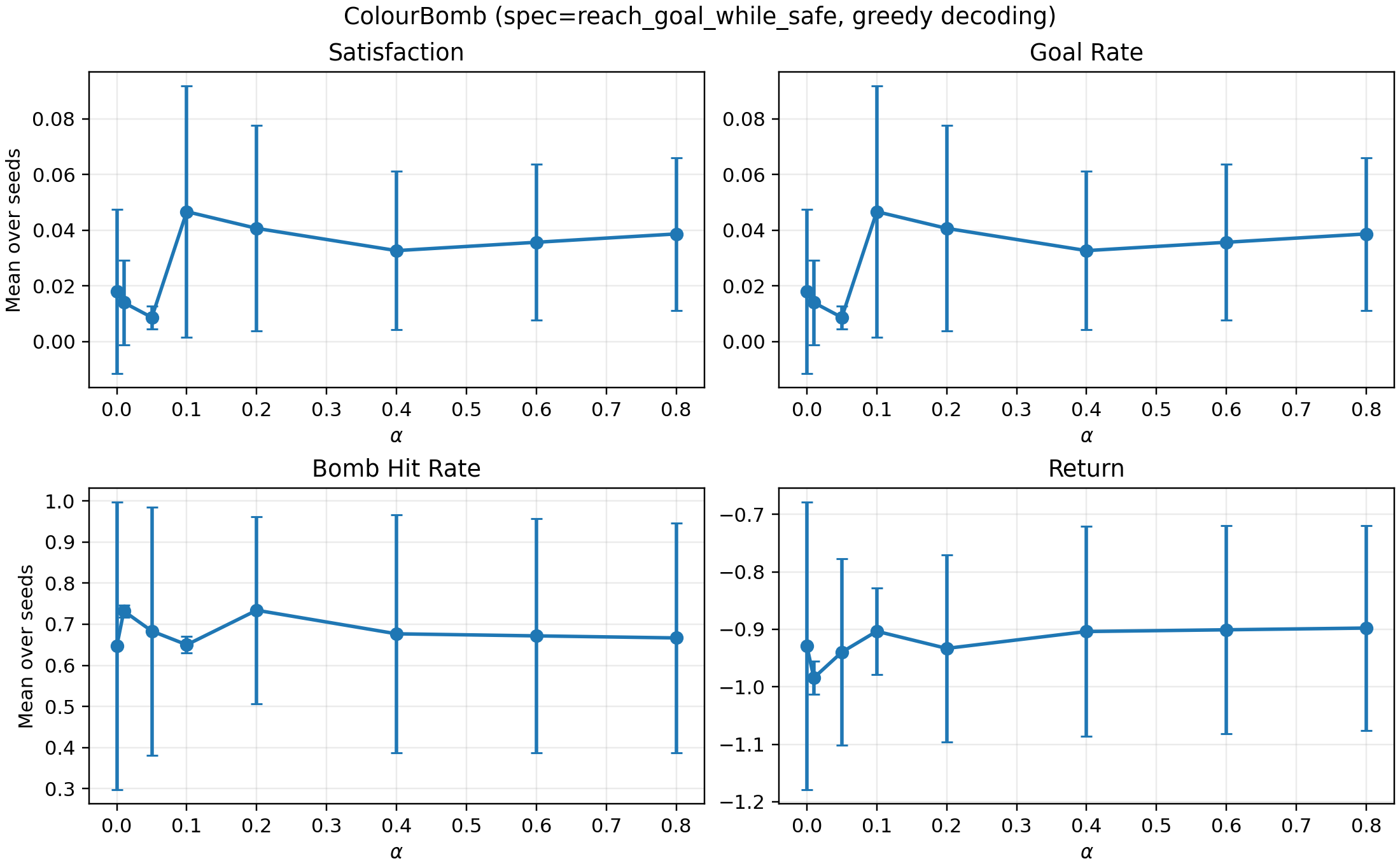}
\caption{Results under the Reach-while-Safe constraint as a function of the weighting loss parameter $\alpha$. Plots report the mean and standard deviation over three runs.}
\label{fig:cb_alpha_tradeoff}
\end{figure}

\paragraph{Reach-while-Safe.}
We next include the goal state in the \ltlf constraint by the specification $G(\neg bomb) \land F(goal)$, which requires the agent to reach a goal while remaining safe throughout the trajectory.
Figure \ref{fig:cb_alpha_tradeoff} summarizes the results for this specification. As expected, satisfying the constraint becomes significantly harder. As a result, the satisfaction rate is substantially lower than in the invariant safety case.
Nevertheless, we can still observe that our approach outperforms the baseline for all values of $\alpha \geq 0.1$, with $\alpha=0.1$ providing the best results. 
This shows that logic regularization still leads to improvements for more complex specifications.
It is worth highlighting that, also in this case, the return associated with values $\alpha \geq 0.1$ is slightly better than what we observe for the baseline.

\paragraph{Cross-Specification Comparison.}
Table \ref{tab:cb_key_results} compares the best configurations for each specification. The invariant safety objective can be improved primarily by reducing bomb-hit events, which directly increases the probability of satisfying the safety constraint. In contrast, the Reach-while-Safe objective remains limited by the difficulty of achieving the goal while maintaining safety simultaneously. In many trajectories, the agent avoids bombs successfully but fails to reach the goal within the episode horizon. This result highlights the intrinsic trade-off between safety and task completion in environments where safe paths towards the goal may be longer or harder to discover.  
%Figure \ref{fig:cb_metrics_bar_example} summarizes results for both \ltlf constraints.

\begin{comment}
\begin{figure}[!th]
\centering
\includegraphics[width=0.75\linewidth]{figures/colourbomb/cb_spec_comparison_goal_vs_safety.png}
\caption{Best-performing value of $\alpha$ for each specification. Results are aggregated over three runs.}
\label{fig:cb_cross_spec}
\end{figure}
\end{comment}

%\begin{figure}[!th]
%\centering
%\includegraphics[width=\linewidth]{figures/colourbomb/cb_metrics_bar_example.png}
%\caption{Summary of performance metrics for both specifications, comparing the vanilla model and logic-regularized variants.}
%\label{fig:cb_metrics_bar_example}
%\end{figure}

\paragraph{Quantitative Summary.}
Table~\ref{tab:cb_key_results} reports representative configurations illustrating the trade-offs observed in the experiments.
For the invariant safety task, the logic-regularized model with $\alpha=0.8$ increases both the satisfaction rate, the goal rate, and the return. For the Reach-while-Safe specification, the best configuration is $\alpha=0.1$, where our approach again improves both the satisfaction rate and the return. It is worth noting that including the reachability of the goal state in the trace constraint helps us in improving the goal-reached rate itself, while slightly degrading performances with respect to the bomb-hit rate.
These results confirm that incorporating logical knowledge into the training objective can improve safety-related metrics while preserving competitive performance in both DT and TT. In both cases, tuning the $\alpha$ parameter remains crucial for achieving high performance. However, compared to DT, the improvements observed in the TT experiments are surprisingly more moderate. We leave to future work a deeper investigation into the reasons behind this difference.

\begin{table}[!t]
\centering
%\small
\begin{tabular}{l l c c c c}
\toprule
Spec & Setting & Satisfaction & Goal & Bomb Hit & Return \\
\midrule
\texttt{safety} & vanilla ($\alpha=0$) & 0.353 & 0.018 & 0.647 & -0.929 \\
\texttt{safety} & logic ($\alpha=0.8$) & \textbf{0.499} & \textbf{0.027} & \textbf{0.470} & \textbf{-0.792} \\
%\texttt{safety} & logic ($\alpha=0.05$) & 0.261 & \textbf{0.049} & 0.739 & -0.946 \\
\midrule
\texttt{reach-while-safe} & vanilla ($\alpha=0$) & 0.018 & 0.018 & \textbf{0.647} & -0.929 \\
\texttt{reach-while-safe} $\,$ & logic ($\alpha=0.1$) $\,$ & \textbf{0.047} $\,$ & \textbf{0.047} $\,$ & 0.650 $\,$ & \textbf{-0.903} \\
\bottomrule
\end{tabular}
\caption{Trajectory Transformer key results.}
\label{tab:cb_key_results}
\end{table}

\section{Conclusion}

We presented a neurosymbolic framework for injecting \ltlf constraints into transformer-based policies for offline RL. Our approach aims at constructing an automaton-aware decision mechanisms that guide trajectory generation, and it relies on two features: a differentiable representation of DFAs and a differentiable logic loss that regularizes the training objective. 
The proposed framework is architecture-agnostic and can be applied to different models like Trajectory Transformers and Decision Transformers. Unlike related popular approaches, we don't need to engineer the underlying reward function, nor we need to focus on a small subset of temporal specifications.
Preliminary experiments on the ColourBomb navigation benchmark show that logic regularization can improve constraint satisfaction while maintaining competitive task performance.
These findings suggest that integrating symbolic background knowledge into offline sequence-based RL is a promising direction for improving policy reliability in safety-critical domains. Future work will extend the empirical evaluation to additional benchmarks, model architectures, temporal specifications, and decoding mechanisms. It would be significant also to perform an extensive end-to-end comparison with well-established frameworks for Safe RL.

\begin{acknowledgments}
This work has been supported by the the PNRR MUR project FAIR (No. PE0000013), and the Italian National Ph.D. on Artificial Intelligence at Sapienza University of Rome. 
\end{acknowledgments}

%% The declaration on generative AI comes in effect
%% in Janary 2025. See also
%% https://ceur-ws.org/GenAI/Policy.html
\section*{Declaration on Generative AI}
%  {\em Either:}\newline
%  The author(s) have not employed any Generative AI tools.
%  \newline

% \noindent{\em Or (by using the activity taxonomy in ceur-ws.org/genai-tax.html):\newline}
During the preparation of this work, the authors used ChatGPT and Claude in order to: Improve writing style. 
%Further, the author(s) used X-AI-IMG for figures 3 and 4 in order to: Generate images. 
After using these services, the authors reviewed and edited the content as needed and take full responsibility for the publication's content. 

%%
%% Define the bibliography file to be used
\bibliography{bibliography}

\end{document}